\documentclass[runningheads]{llncs}

\usepackage{eccv}
\usepackage[pagebackref,breaklinks,colorlinks,citecolor=eccvblue]{hyperref}
\usepackage{orcidlink}

\newcommand{\method}{EgoPoser\xspace}

\def\SLOW{\textsc{SLOW}}
\def\FAST{\textsc{FAST}}

\def\SMPLbeta{\mathbf{\beta}}

\usepackage{xifthen}

\newcommand{\norm}[1]{\left\lVert#1\right\rVert}

\newcommand{\normal}[3][]{%
    \ifthenelse{\isempty{#1}}
        {\mathcal{N}\left(#2, #3\right)}
        {\mathcal{N}\left(#1|#2, #3\right)}%
}

\usepackage[ruled]{algorithm2e} 

\SetAlFnt{\small}
\SetAlCapFnt{\small}
\SetAlCapNameFnt{\small}
\SetAlCapHSkip{0pt}

\usepackage{epsfig}
\usepackage{graphicx}
\usepackage{amsmath}

\usepackage{multirow}
\usepackage{tabularx}
\usepackage{multicol}
\usepackage{booktabs}
\usepackage{gensymb}
\usepackage{array}
\usepackage{comment}
\usepackage{color}
\usepackage{adjustbox}
\usepackage{xcolor}
\usepackage{xspace}
\usepackage{todonotes}
\usepackage{wasysym}
\usepackage{ragged2e}
\usepackage{environ}
\usepackage{wrapfig}

\usepackage{overpic}

\usepackage[accsupp]{axessibility}  

\definecolor{myblue}{RGB}{0,204,255}
\definecolor{mygray}{RGB}{192,192,192}
\definecolor{myorange}{RGB}{255,127,0}
\definecolor{mygreen}{RGB}{5,102,8}
\definecolor{mypink}{RGB}{231,84,128}
\definecolor{myred}{RGB}{255,0,0}

\begin{document}
\title{\method: Robust Real-Time Egocentric Pose Estimation from Sparse and Intermittent Observations Everywhere}

\author{Jiaxi Jiang,\quad
Paul Streli,\quad
Manuel Meier,\quad
Christian Holz\\
}
\titlerunning{EgoPoser: Robust Real-Time Egocentric Pose Estimation}
\authorrunning{Jiang et al.}
\institute{ 
Department of Computer Science, ETH Zürich, Switzerland 
\\
{\tt\small \{firstname.lastname\}@inf.ethz.ch}\\
\url{https://siplab.org/projects/EgoPoser}
}

\maketitle

\begin{abstract}
Full-body egocentric pose estimation from head and hand poses alone has become an active area of research to power articulate avatar representations on headset-based platforms.
However, existing methods over-rely on the indoor motion-capture spaces in which datasets were recorded, while simultaneously assuming continuous joint motion capture and uniform body dimensions.
We propose \method to overcome these limitations with four main contributions.
1)~\method robustly models body pose from intermittent hand position and orientation tracking only when inside a headset's field of view.
2)~We rethink input representations for headset-based ego-pose estimation and introduce a novel global motion decomposition method that predicts full-body pose independent of global positions.
3)~We enhance pose estimation by capturing longer motion time series through an efficient SlowFast module design that maintains computational efficiency.
4)~\method generalizes across various body shapes for different users.
We experimentally evaluate our method and show that it outperforms state-of-the-art methods both qualitatively and quantitatively while maintaining a high inference speed of over 600\,fps.
\method establishes a robust baseline for future work where full-body pose estimation no longer needs to rely on outside-in capture and can scale to large-scale and unseen environments.

\end{abstract}

\begin{figure}
\includegraphics[width=\textwidth]{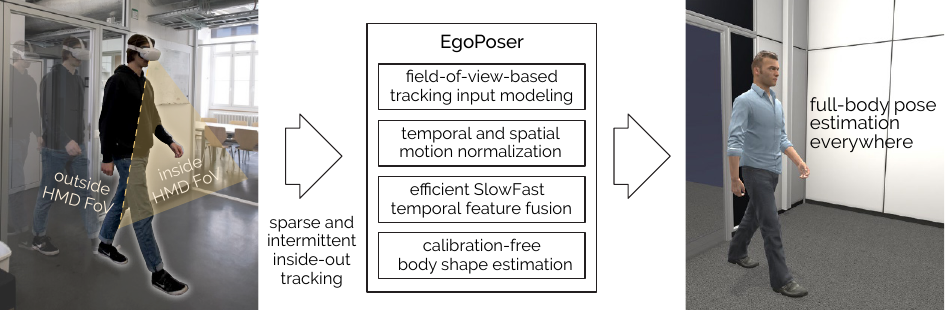}
\caption{Today's Mixed Reality systems integrate all tracking inside the headset, supporting mobile use in everyday environments.
This sacrifices much of the user's body and surroundings for input, when body parts leave the cameras' field of view.
Accounting for these constraints, our novel method \method robustly estimates full-body poses that are spatially and temporally coherent, even from the sparse and intermittent inside-out tracking input available on today's headsets.
}
\label{fig:teaser}
\end{figure}

\section{Introduction}

Current Mixed Reality (MR) systems such as Microsoft HoloLens, Meta Quest, and Apple Vision Pro derive tracking cues and user input mainly from a head-worn platform.
The cameras inside these devices observe the environment as well as the user's hand motions when inside the field of view (FoV)~\cite{han2020megatrack, han2022umetrack}.
This enables the system to track the device's position inside the world and derive input commands from the user's actions.
Due to the sparse nature of the input signals, whose capture relies on data from the user's head and hands, current MR systems are limited in their ability to generate comprehensive virtual avatar representations, often limited to only their upper body.
This reduces the fidelity of the user experience and may also affect the user's sense of immersion.

Motivated by the goal to holistically embody users as full-body avatars in MR, several recent methods have been developed to estimate full-body poses from the sparse tracking cues current systems provide~\cite{jiang2022avatarposer,aliakbarian2022flag,yang2021lobstr,dittadi2021full, jiang2024manikin, ponton2023sparseposer}.
These efforts all rely on large motion-capture datasets to estimate realistic body poses and animations, leveraging the robust, continuous, and high-fidelity recordings across a large variety of environments.

However, existing methods exhibit several limitations in real-world applications. 
(1)~Prior approaches directly use the global pose in world space as the network input, causing the trained model to overfit to motions and poses that are specific to the environment and typically concentrated near the origin. 
Our paper reveals that using global input representation results in significantly worse predictions, even for slight meter-scale offsets.
(2)~Current methods assume that the pose of users' hands is always available, which would require them to always remain within the field of view of the headset's cameras. 
However, portable inside-out tracking systems obtain only intermittent information about the hands, as they occasionally move out of the field of view. 
(3)~Existing methods only account for a mean body shape and disregard the natural variations in body shapes across different people.
This limitation prevents the model from adapting to real-world inputs and accurately representing the user's body. 
Often, motion artifacts such as floating and ground penetration arise from this.

To address these problems, we propose \method, an exclusively headset-centered estimation method for full-body poses that robustly performs on the sparse and intermittent tracking cues provided by today's inside-out tracking systems.
As shown in Fig.~\ref{fig:teaser}, \method\ comprises four main components that jointly enable its robust performance on real-world data and live motions outside motion-capture datasets.
(1)~\method's realistic field-of-view modeling captures both spatial and temporal information to smoothly estimate accurate full-body poses even when the user's hands leave the camera's view frustum.
(2)~Our novel global motion decomposition retains the critical relative global information using local representation, making it robust to position changes by encoding motion priors from sparse inputs.
(3)~We sample the original signals at different rates to capture longer motion time series through a SlowFast module as the input to the Transformer encoder, thus improving prediction accuracy without increasing the computational burden.
(4)~To support personalized use, we predict individual body shapes to accurately anchor each user's representation within the virtual environment.

Taken together, we make the following contributions in this paper:

(1)~We propose \method, a novel systematic approach to full-body pose estimation from the signals HMD devices provide. \method remains robust even when hands leave the field of view, and it generalizes well to various body shapes.

(2)~We have identified a notable issue with existing methods: they tend to overfit to the training data due to the global input representation of the neural network. We emphasize the significance of position-invariant prediction and present an effective global motion decomposition strategy in \method.

(3)~\method effectively accommodates to different body shapes unlike previous methods. We demonstrate our method's input adaptability and the accurate output avatar representation it produces.
In addition, \method significantly reduces motion artifacts such as floating and ground penetration. 

(4)~We demonstrate superior numerical and visual performance compared to state-of-the-art methods on the public datasets AMASS and HPS. Our demo also shows that \method can operate with real-world MR systems, making it a practical and effective solution for use with end-user devices.

\section{Related Work}

\subsubsection{Egocentric human pose estimation for Mixed Reality.}
The task of estimating full-body poses from head and hand poses alone has gained significant attention with the growing popularity of MR-based egocentric vision~\cite{grauman2022ego4d, grauman2024ego, ma2024nymeria}. Our previous work AvatarPoser~\cite{jiang2022avatarposer} trained a single model for various motion types, combining a Transformer-based neural network with inverse kinematics (IK) optimization for realistic predictions that match the observations.
QuestSim~\cite{winkler2022questsim} and QuestEnvSim~\cite{lee2023questenvsim} combined reinforcement learning with physics simulation to ensure physically plausible predictions. 
While diffusion model-based methods such as AGRoL~\cite{du2023avatars} and EgoEgo~\cite{li2023ego} synthesized smooth predictions, they both relied on future input signals to predict the current frame, and supported only slow sampling speeds---two factors that pose significant challenges for real-time applications.
Recently, a temporal-spatial Transformer-based method~\cite{zheng2023realistic} was proposed that can produce realistic results by joint-level modeling, but its network design is still computationally expensive.
Our recent work MANIKIN~\cite{jiang2024manikin} introduces a biomechanically accurate, differentiable full-body IK framework that matches observations and reduces ground penetration.
However, as we demonstrate in this paper, these previous projects~\cite{jiang2022avatarposer, du2023avatars,li2023ego,zheng2023realistic, jiang2024manikin} take the global pose as the network input, which is prone to overfitting to the training data and cannot generalize well to unseen positions.
The design of \method thus aims to robustly perform independent of these positions.

\subsubsection{Human pose estimation from inertial sensors.}
In addition to MR devices, there has also been research on full-body pose estimation from body-worn inertial sensors~\cite{von2017sparse, DIP:SIGGRAPHAsia:2018, yi2021transpose, yi2022physical, jiang2022transformer, armani2024ultra, yi2023egolocate, mollyn2023imuposer, yi2024physical, van2024diffusionposer}.
Because these sensors are distributed across the body, motion capture (MoCap) can become inflexible and obtrusive.
Besides, inertial pose estimation usually struggles with accurate global position tracking since only relative position and orientation changes can be observed as input.
In contrast, our task uses the three-point tracking of today's MR devices without additional instrumentation. Without trackers on the lower body, we infer the complete body pose, including leg motions, from the global movements of the head and hands.

\subsubsection{Pose estimation under field-of-view constraints.}
Estimating human pose when parts of the body are outside the cameras' field of view is challenging~\cite{ahuja2019mecap, rhodin2016egocap, wu2020back, bailly2012shoesense, parger2021unoc,xie2023visibility}. 
To alleviate the visibility issue, previous work has investigated alternative configurations with custom sensors such as wrist-worn cameras~\cite{li2020mobile} or IMUs~\cite{streli2023hoov}, cameras embedded inside controllers~\cite{ahuja2022controllerpose}, and hat-mounted downward-facing fisheye cameras~\cite{rhodin2016egocap, wang2021estimating,akada2022unrealego} or glasses~\cite{zhao2021egoglass, kang2023ego3dpose}.
In terms of software solutions, FLAG~\cite{aliakbarian2022flag} retained the original constraints of headset-only capture and augmented the training data by randomly masking the hands with a certain probability instead.
While promising, it did not consider the spatial relative pose between the hand and headset, as hands can be masked out even if they are actually inside the FoV.
In this paper, we realistically model the cameras' FoV by considering the spatial relative pose and the temporal continuity.

\section{Proposed Method: \method}

We now describe our method \method for the real-time estimation of the global full-body pose based on head and hand poses tracked by an HMD device from egocentric vision.
We first outline the specific task we investigate and then describe our method.

\subsection{Overview}
While MR systems differ in the tracking technology they implement, they all typically provide the global positions $\mathbf{p}$ and orientations $\mathbf{\Theta}$ of the headset, along with those of the user's two hands.
Following AvatarPoser~\cite{jiang2022avatarposer}, we aim to find a mapping $f$ from the 3-point pose input to the user's full-body pose, and thus the positions of all $J$ full-body joints:

\begin{equation}
    \{\mathbf{p}^j_t\}^{j=1:J} = f(\{\mathbf{p}^j_t, \mathbf{\Theta}^j_t\}^{j=1:3})
\end{equation}
This is a challenging under-determined problem because the same input may correspond to multiple possible outputs. To ensure the prediction of consistent human skeletons, prior work used the first 22 joints defined in the kinematic tree of the SMPL-H~\cite{loper2015smpl,AMASS:ICCV:2019} skeleton model as the output representation of the full body, ignoring the pose of the fingers. Additionally, the SMPL-H model parameterizes the shape of the corresponding 3D human mesh with 16 shape parameters, $\mathbf{\beta} \in \mathbb{R}^{16}$.

\begin{figure*}[t]
    \centering
    \includegraphics[width=1\linewidth,trim={0 0mm 0 0mm},clip]{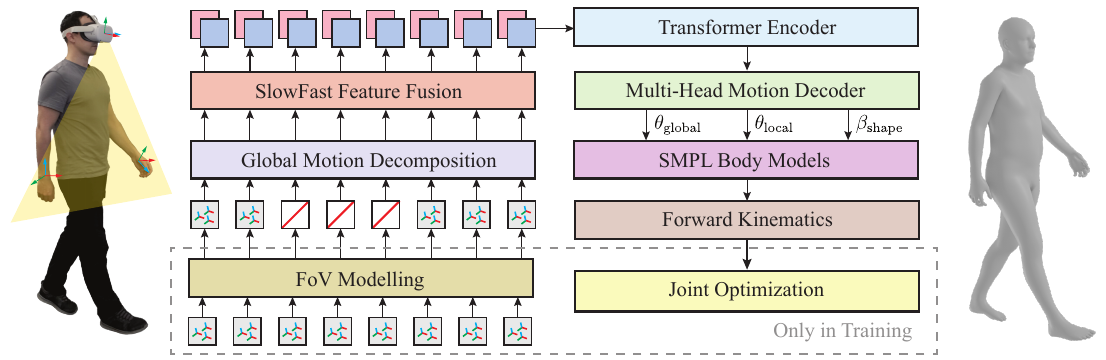}
    \caption{The architecture of \method\ for full-body pose estimation from an MR device. 
    Given N=80 frames as input, we generate the last frame as the full-body representation for each timestamp, facilitating real-time applications.
    }
    \label{fig:arch}
\end{figure*}

We show an overview of our approach in Fig.~\ref{fig:arch}. The core components of our method include our proposed realistic FoV modeling, global motion decomposition, SlowFast feature fusion module, Transformer Encoder, and a human motion decoder.
The output of the human motion decoder consists of the global root orientation $\mathbf{\theta}_\text{global}$, the local joint rotations $\mathbf{\theta}_\text{local}$, and the shape parameters $\SMPLbeta$. The pose and shape parameters are then jointly optimized using the SMPL body model and forward kinematics. As proposed in AvatarPoser~\cite{jiang2022avatarposer}, the global root position is calculated using forward kinematics based on the tracked head position and the predicted joint angles.

\subsection{Realistic Field of View Modeling}

Previous work does not adequately address the inherent limitations of the inside-out hand tracking on today's state-of-the-art headsets such as Apple Vision Pro, Meta Quest~2/3/Pro, and HoloLens~2.
They typically model tracking failures through random frame drops uniformly sampled over the complete tracked motion but fail to account for the fact that hand tracking generally fails in regions outside the headset cameras' field of view.

\begin{figure}[]
    \centering
    \includegraphics[width=\linewidth]{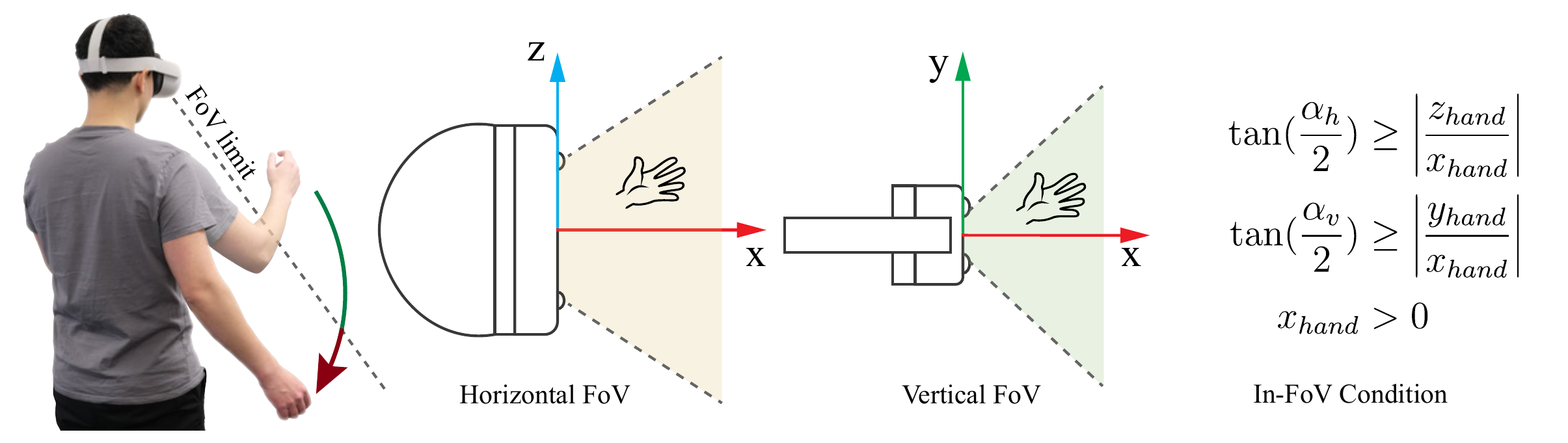}
    \caption{An illustration of an HMD's field of view and in-FoV conditions.}
    \label{fig:fov}
\end{figure}

Based on the head pose, which determines the viewing angle of the cameras mounted on the headset, and the relative position of the hands, we simulate hand tracking failures for headsets with varying FoVs.

Fig.~\ref{fig:fov} shows an HMD's field of view and the corresponding in-FoV conditions, where $\alpha_h$ is the horizontal FoV and $\alpha_v$ is the vertical FoV. 
Here, $x_{\text{hand}}$, $y_{\text{hand}}$, and $z_{\text{hand}}$ are the x, y, and z coordinates of the hand position in a head-centered coordinate system with the x-axis pointing through the eyes. 
We train our method to robustly handle inputs with continuous tracking gaps by setting the input features of joints outside the field of view to $0$. 

\subsection{Global Motion Decomposition}
Today's MR headsets track their own global pose as well as the user's hands or controllers in three-dimensional space.
Based on that, previous methods~\cite{jiang2022avatarposer, du2023avatars, zheng2023realistic} use the global poses in world space as input to their networks. However, most existing datasets, like AMASS, are recorded in a limited physical space near the origin, so it remains unclear whether these methods can generalize well to different locations. Data augmentation could be a remedy, yet it makes the training process less efficient and it is impossible to cover the entire infinite 3-space. 
One common strategy is to decompose the global motion into a rigid body motion in global world space and a local motion relative to a root capturing the current body pose.
However, in 3-point tracking problems, simply converting the reference frame to the head, the root frame, makes the prediction sensitive to head rotation. Additionally, removing the global information can lead to information loss, making the ill-posed problem even more challenging.

\begin{figure}
    \begin{minipage}{.48\textwidth}
    \centering
    \includegraphics[width=\linewidth]{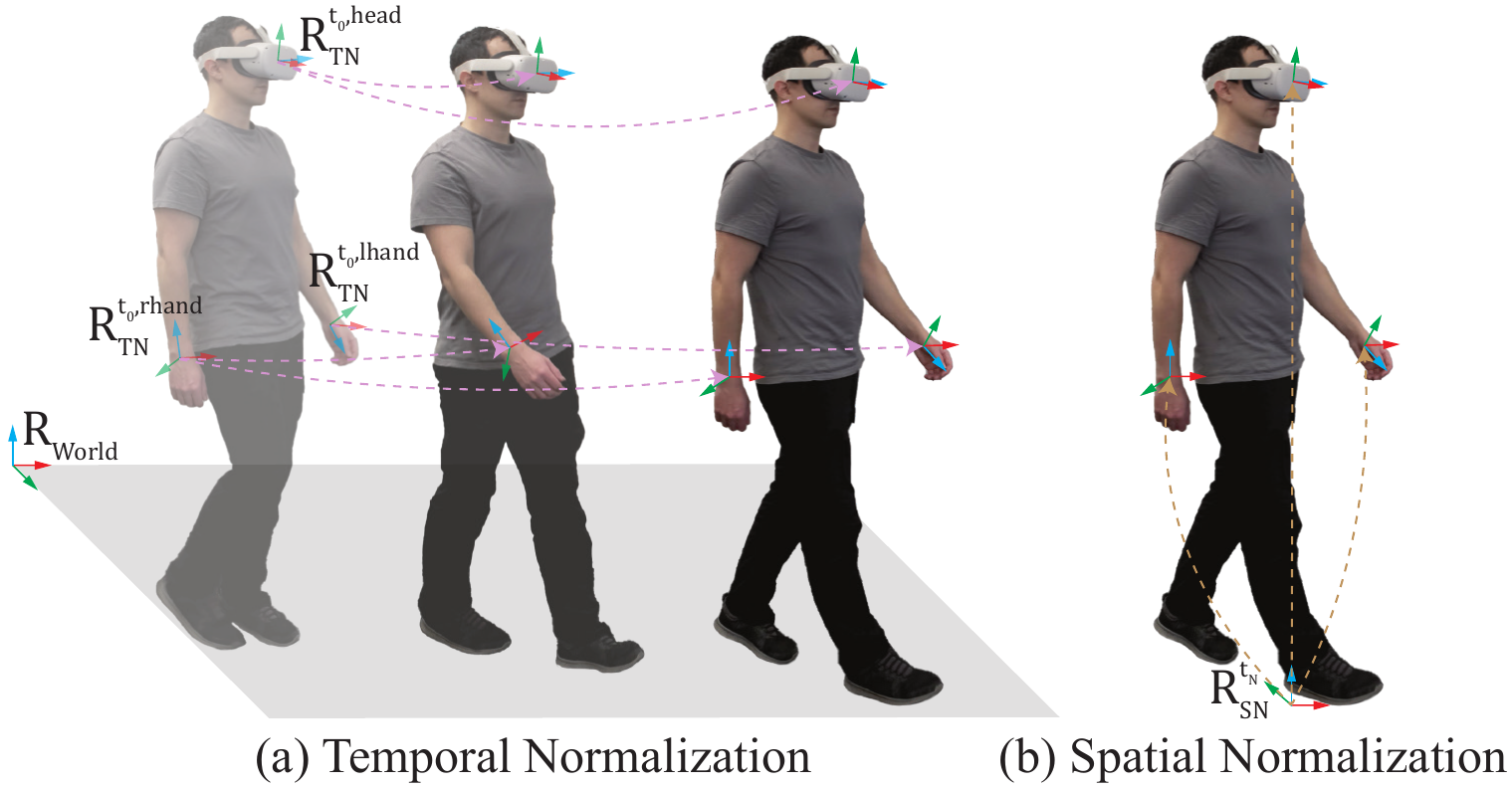}
    \caption{An illustration of the temporal and spatial normalizations for robust position-invariant pose estimation.}
    \label{fig:normalization}
        \end{minipage}
    \hfill
    \begin{minipage}{.48\textwidth}
    \centering
    \includegraphics[width=\linewidth]{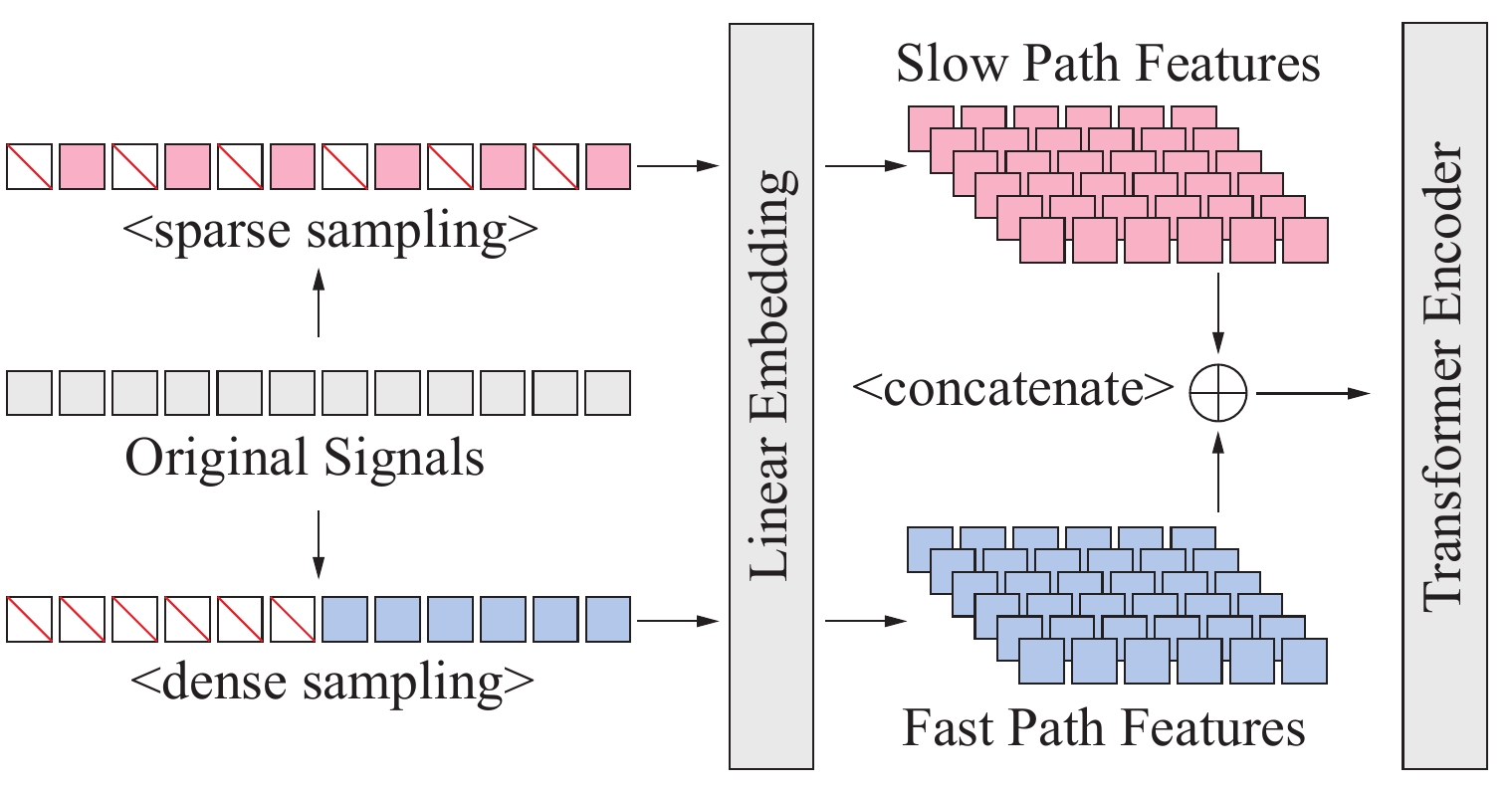}
    \caption{SlowFast feature fusion module. Original signals are sparsely and densely sampled and then concatenated.}
    \label{fig:slowfast}
        \end{minipage}
\end{figure}


To combine the advantages of both global and local representations, we introduce a global motion decomposition strategy designed to be position-invariant for pose estimation across large-scale environments. 
Our approach contains two key operations, temporal and spatial normalization, detailed as follows:\\

\noindent\textbf{(1) Temporal Normalization (TN).} We perform temporal normalization by subtracting the translation of each joint at the first frame from the corresponding joint positions over the temporal window. This operation extracts the relative global trajectory of each joint across the temporal window.\\

\noindent\textbf{(2) Spatial Normalization (SN).} Instead of subtracting the head's 3D translation from both the head and hand poses to obtain head-relative hand positions, we normalize only the horizontal translations relative to the head. The global vertical translation is retained as a crucial feature to encode motion priors.

Fig.~\ref{fig:normalization} illustrates the spatial and temporal normalizations. These normalizations translate the positions of the headset and hands from a world reference frame $R_{\text{World}}$ anchored at a static ground location to a reference frame $R_{\text{SN}}^{t}$ anchored at the head's projected ground location, and a reference frame $R_{\text{TN}}^{t,j}$ anchored at the $j$-th joint's initial position for a given window. These transformed positions are then used as input to our pose estimation network. The positions provided by SN and TN at time $t_i$ are written as:

\begin{equation}
\begin{aligned}
\mathbf{p}_{\text{SN},h}^{t_i,\text{hand}} &= \mathbf{p}_{W,h}^{t_i,\text{hand}} - \mathbf{p}_{W,h}^{t_i,\text{head}}\\
\mathbf{p}_{\text{TN}}^{t_i,j} &= \mathbf{p}_{W}^{t_i,j} - \mathbf{p}_{W}^{t_0,j}
\end{aligned}
\end{equation}
In addition to the orientation and decomposed position information, we calculate the corresponding linear and angular velocities of the head and hands to enrich the input data. For the rotations, we use 6D representations~\cite{zhou2019continuity} to ensure continuity. Finally, a total of 59 input features are provided to the network.

\subsection{SlowFast Feature Fusion}

Based on the information from a single frame, a multitude of plausible body poses exist that would fit a given set of head and hand poses.
However, the problem converges toward a more unique solution as we consider the head and hand motions over a longer temporal context.
Yet, simply adding more frames to the input would significantly increase computational overhead.
For example, the computational complexity of a Transformer's self-attention module scales quadratically with input sequence length. Thus, inspired by SlowFast networks originally proposed for video recognition~\cite{feichtenhofer2019slowfast}, we propose a SlowFast feature fusion module that increases the context of considered past tracking frames in a more efficient manner.

As shown in Fig.~\ref{fig:slowfast}, given an input window of $\tau$ past frames, the SlowFast module concatenates the linear embeddings for the last $\frac{\tau}{2}$ frames (\FAST{}) with $\frac{\tau}{2}$ frames sampled with a stride of 2 over the complete window (\SLOW{}).
In this way, we reduce the length of the input sequence by a factor of 2 while maintaining the temporal context over the whole window.
Additionally, we still capture the temporal information contained within the higher temporal resolution of the \FAST{} input frames.

\subsection{Shape-aware Pose Optimization}
\label{sec:calibbody}
One common limitation of existing methods~\cite{jiang2022avatarposer, du2023avatars, dittadi2021full} is that they assume a mean-shaped body skeleton and mesh, ignoring differences in body sizes across users. This assumption introduces two problems: First, since all the training and testing data share the same body shape, it is unclear whether a trained model will generalize well to input data from users with diverse body shapes in real-world scenarios. Second, the animation based on the mean-shaped skeleton does not accurately reflect the user's real body size. Even with perfect joint angles, this can result in ground penetration, self-penetration, or floating artifacts.\\

\noindent\textbf{Solution 1: Data augmentation + calibration.}
To address this issue, we introduce an approach that combines data augmentation with T-pose calibration. We augment the training data by incorporating ground truth shape parameters. For evaluation on the test set, we measure the body height and arm length and compute the ratio between the measured and the corresponding mean shape values. The average of these ratios is then used as the scaling factor for the entire skeleton and applied to adjust the output representation accordingly.\\

\noindent\textbf{Solution 2: Joint pose and shape estimation.} 
The previous solution does not account for variations in body ratios and requires an extra calibration step to approximate a user's body size, adding to the overall effort involved. To address this, we introduce a calibration-free method that estimates the user's shape from the tracking input.

Instead of directly supervising the shape parameters $\SMPLbeta$ in the loss function, we implicitly optimize the estimated $\SMPLbeta$ through the error in the joint positions generated by the shape-aware differentiable SMPL body model, which takes $\SMPLbeta$ and the estimated joint rotations as input. Additionally, we apply L1 regularization to $\SMPLbeta$ to encourage sparsity, ensuring that shape parameters which do not impact joint locations and bone lengths default to zero.

As our method estimates $\SMPLbeta$ for each frame, we can apply the median shape parameters from an initial sequence of frames to enforce consistency. However, we did not observe frequent or sudden deviations in $\SMPLbeta$ for a given input sequence.

\noindent\textbf{Loss functions.} The loss function for body shape estimation is written as: 
\begin{equation}
\begin{aligned}
    \mathcal{L}_\mathrm{shape} &= \mathcal{\lambda}_{\mathrm{pos}} \mathcal{L}_{\mathrm{pos}}+ \mathcal{\lambda}_{\mathbf{\beta}} \norm{\SMPLbeta}_{1}
\end{aligned}
\end{equation}
where the shape-guided positional loss is calculated through forward kinematics:
\begin{equation}
 \mathcal{L}_{\mathrm{pos}}=\norm{\mathrm{FK}(\mathbf{\theta}, \mathbf{\beta}) - \mathrm{FK}(\mathbf{\theta}_{GT}, \mathbf{\beta}_{GT})}_1 \\ 
\end{equation}
The final loss function is composed of an L1 global orientation loss, an L1 local rotational loss, an L1 positional loss, and an L1 regularization of $\SMPLbeta$ denoted by:
\begin{equation}
    \mathcal{L}_{\mathrm{total}} = \mathcal{\lambda}_{\mathrm{ori}} \mathcal{L}_{\mathrm{ori}} + \mathcal{\lambda}_{\mathrm{rot}} \mathcal{L}_{\mathrm{rot}} + \mathcal{\lambda}_{\mathrm{pos}} \mathcal{L}_{\mathrm{pos}} + \mathcal{\lambda}_{\mathbf{\beta}} \norm{\SMPLbeta}_{1}
\end{equation}
We set the weights $\mathcal{\lambda}_{\mathrm{ori}}$, $\mathcal{\lambda}_{\mathrm{rot}}$, $\mathcal{\lambda}_{\mathrm{pos}}$, and $\mathcal{\lambda}_{\mathbf{\beta}}$ to 0.05, 1, 1, and 0.01 respectively.

\section{Experiments}

\subsection{Datasets and Training Details}

Following prior work~\cite{jiang2022avatarposer}, we used three subsets of the AMASS~\cite{AMASS:ICCV:2019} dataset, namely CMU~\cite{cmu}, BMLrub~\cite{troje2002decomposing}, and HDM05~\cite{cg-2007-2}, for both training and testing. We applied the data split provided by AvatarPoser~\cite{jiang2022avatarposer}, which randomly allocates 90\% of the sequences to the training set and 10\% to the test set. To evaluate performance in the wild, we also used the HPS dataset~\cite{guzov2021human} for testing, which captures subjects within large-scale scenes. 
For ground truth, we relied on the high-quality joint optimization results described by Guzov et al.~\cite{guzov2021human}, who consider camera localization, IMU pose estimates, and scene constraints.

To train \method, we adopted the Adam solver~\cite{kingma2015adam} with a batch size of 256. We considered the latest 80 frames as input ($\tau=80$), resulting in an input window with 40 frames after SlowFast fusion. The learning rate starts from $1\times10^{-4}$ and decays by a factor of 0.5 every $2\times10^4$ iterations. We trained EgoPoser using PyTorch on one NVIDIA GeForce GTX 3090 GPU.

\subsection{Evaluation Metrics}
We use Mean Per Joint Position Error (MPJPE [cm]) and Mean Per Joint Velocity Error (MPJVE [cm/s]) as our main evaluation metrics to determine estimation accuracy and smoothness. To evaluate the shape-aware pose estimation, we use Mean Vertex Error (MVE [cm]), as well as the mean errors of the predicted heights (in [cm]) and bone lengths (in [cm]). In addition, we compute the average distance to the ground for mesh vertices below the ground to evaluate ground penetration~\cite{yuan2021simpoe}. To analyze foot floating artifacts, we calculate the mean distance (in [cm]) between the ground and the lowest vertex of the mesh across all frames where all vertices of the mesh are above the ground.

\begin{figure}[]
    \centering
    \begin{minipage}{.48\textwidth}
    \includegraphics[width=\linewidth]{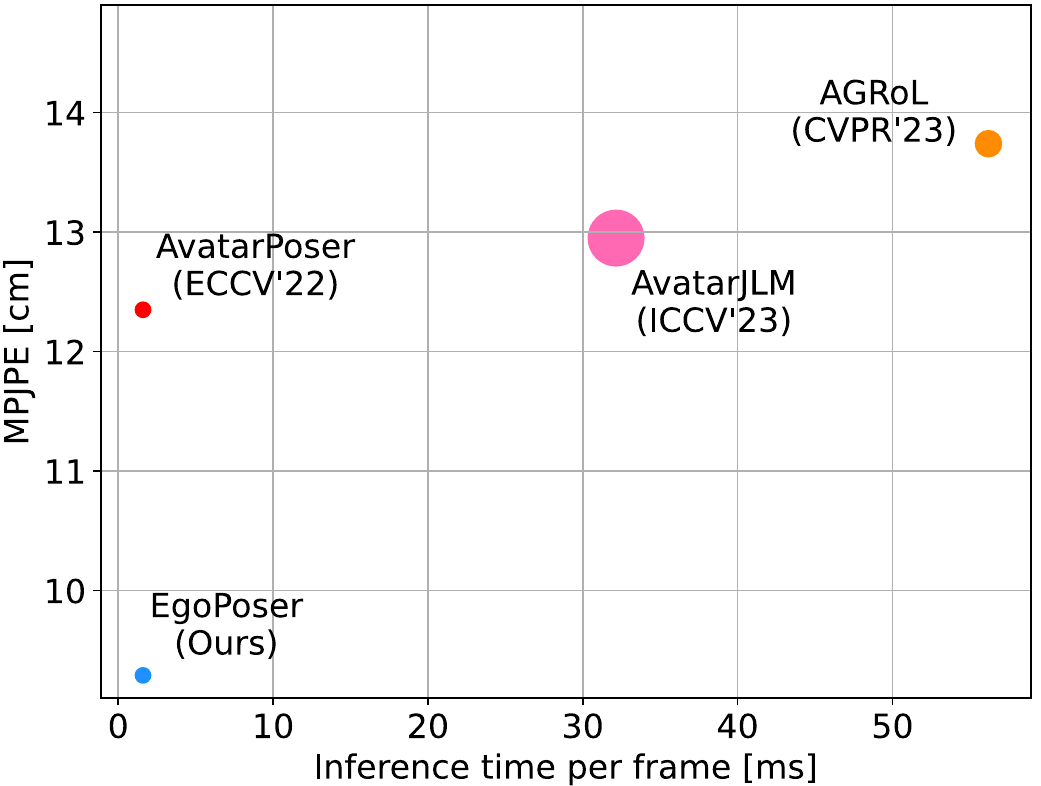}
    \caption{Comparisons of computational complexity. Marker size indicates parameter counts. We achieved the smallest error on the HPS dataset with fast inference.}
    \label{fig:flops}
    \end{minipage}
    \hfill
    \begin{minipage}{.48\textwidth}
    \centering
    \includegraphics[width=\linewidth]{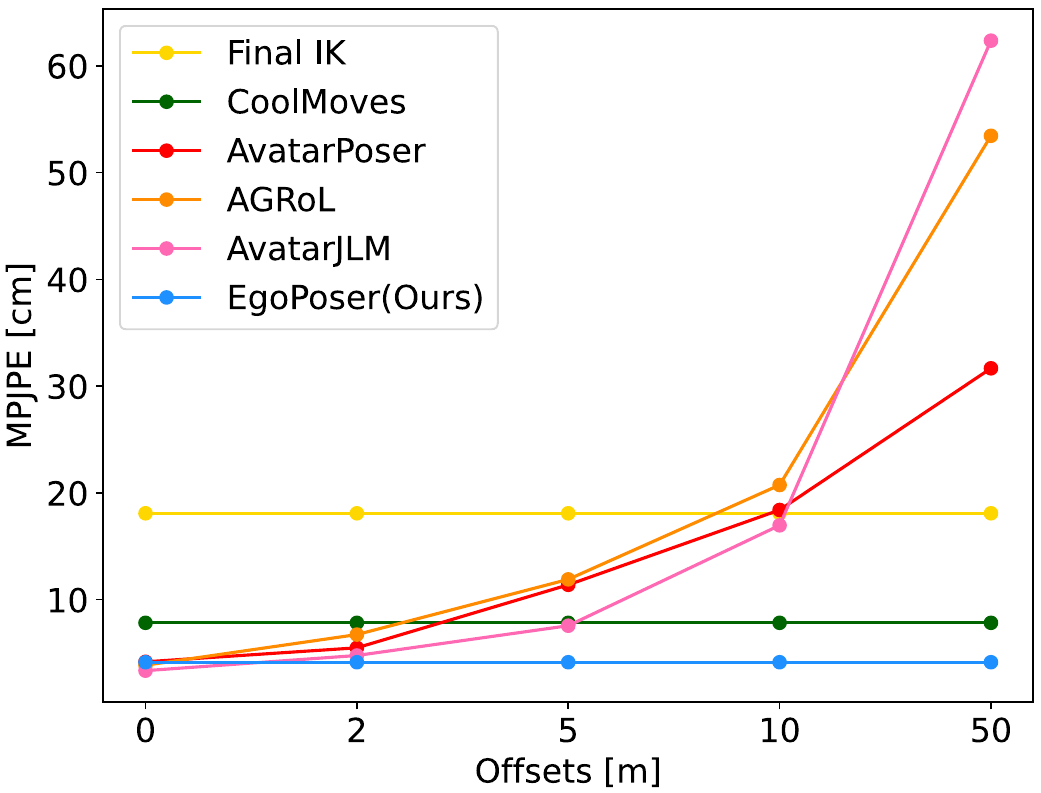}
    \caption{Position error relative to users' distance from the origin. Our method remains robust with MPJPE of 4.14\,cm and MPJVE of 25.95\,cm/s.}
    \label{fig:result_amass}
    \end{minipage}
\end{figure}

To ensure a fair comparison with state-of-the-art methods and to demonstrate the impact of each proposed component, we assume full hand visibility and use the mean body shape for comparisons with prior work making similar assumptions (Tab.~\ref{tab:result_hps} and Fig.~\ref{fig:result_amass}). We evaluate the hand partial visibility problems and size-aware pose estimation independently in Tab.~\ref{tab:fov} and Tab.~\ref{tab:bodysize}, respectively. Our motion decomposition method is employed in all evaluations of EgoPoser to ensure robust performance. This results in a slight increase in positional error when trained and tested on the three AMASS subsets, CMU, BMLrub, and HDM05, which capture motion sequences close to a fixed origin.

\begin{table*}[]
    \centering
    \caption{Comparisons to state-of-the-art methods on HPS dataset, which is captured in large scenes. 
    }
    \setlength{\tabcolsep}{4pt}
    \begin{adjustbox}{width=\textwidth}
    \begin{tabular}{l|cc|cc|cc|cc|cc}
    \toprule
                 & \multicolumn{2}{c}{$\text{BIB\_EG\_Tour}$}&
                \multicolumn{2}{c}{MPI\_EG}&     
                \multicolumn{2}{c}{Working\_Standing}
                & \multicolumn{2}{c}{UG\_Computers}&
                \multicolumn{2}{c}{Go\_Around}\\
         Methods &MPJPE&MPJVE&MPJPE&MPJVE&MPJPE&MPJVE&MPJPE&MPJVE&MPJPE&MPJVE\\
    \midrule
         AvatarPoser~\cite{jiang2022avatarposer} &22.53  & 60.25 & 16.54   &  36.39 & 19.08 & 52.95
         &23.24  & 40.65 & 19.50   &  59.54 \\
         AvatarPoser-Improved &11.48& 82.70& 13.86&  59.66& 12.42& 77.83&11.42& 50.46& 12.56&  82.42\\
         AGRoL~\cite{du2023avatars} & 28.95 & 166.34 &19.41 & 55.52 & 17.67 & 53.97&20.90&109.12&14.16&98.34\\
         AGRoL-Improved & 15.04& 124.12&13.94& 89.42& 13.86& 89.42&12.71&106.43&13.13&128.42\\

 AvatarJLM~\cite{zheng2023realistic} &41.27 &82.92 &12.91 &50.44 &17.26 &69.08 &21.31 &55.42 &11.57 &62.18 \\
  AvatarJLM-Improved &14.80&79.66&14.72&45.57&13.75&68.98&10.28&45.74&11.19
&68.87\\

         EgoPoser (Ours) &  \textbf{9.55}  & \textbf{49.39}  &  \textbf{11.05} & \textbf{35.60}   & \textbf{8.70} & \textbf{46.49} 
         &  \textbf{10.25}& \textbf{38.29}  &  \textbf{6.90} & \textbf{45.10}   \\
    \bottomrule
    \end{tabular}
    \end{adjustbox}
    \label{tab:result_hps}
\end{table*}

\begin{figure}
\centering
\includegraphics[width=\linewidth]{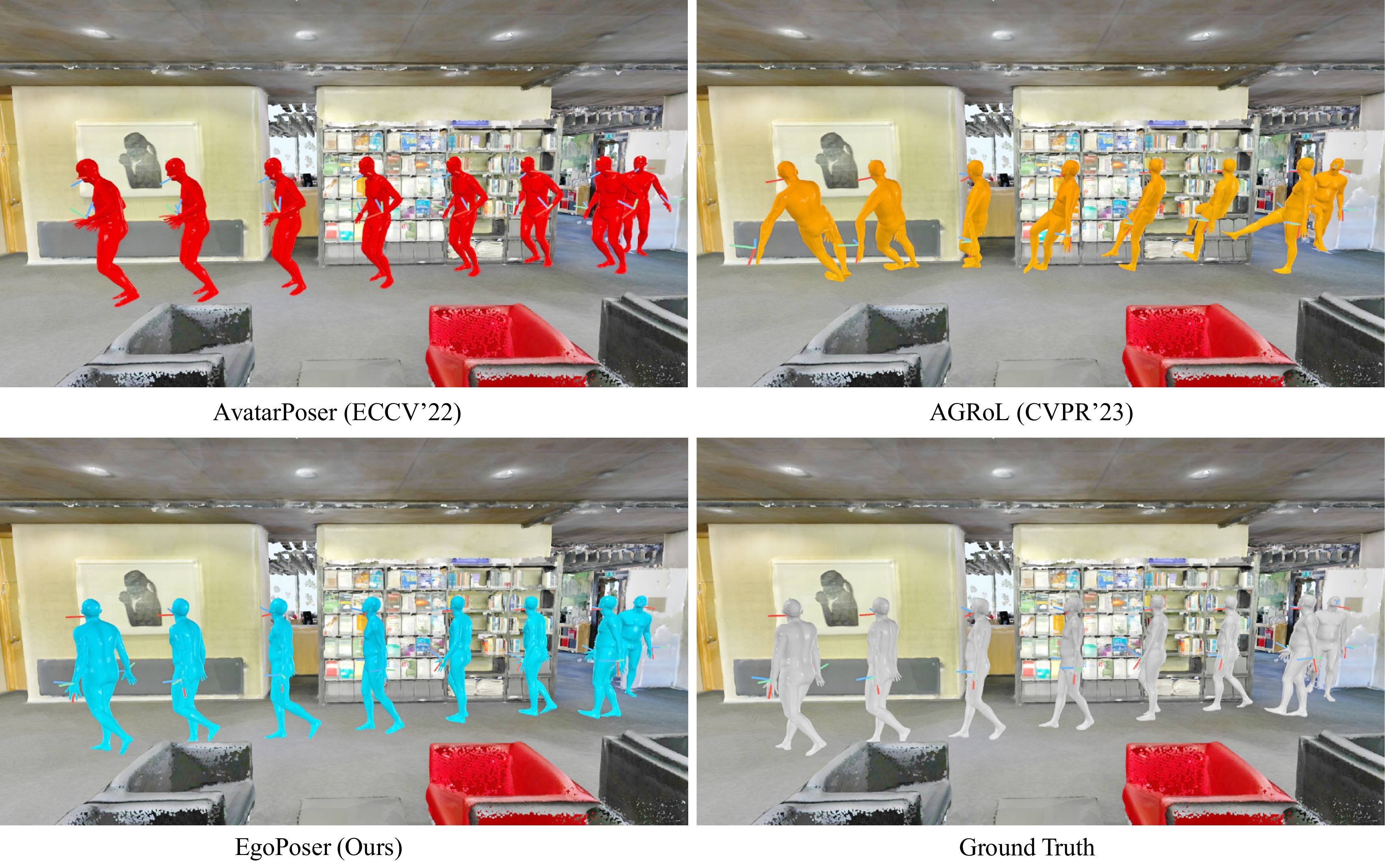}
    \caption{Visual comparisons on the HPS dataset.}
    \label{fig:hps}
\end{figure}

\subsection{Evaluation Results}
\label{sec:inthewild}

\noindent\textbf{Results of pose estimation in the wild.} 
We compare our method to state-of-the-art approaches, including AvatarPoser~\cite{jiang2022avatarposer}, AGRoL~\cite{du2023avatars}, and AvatarJLM~\cite{zheng2023realistic}, on the real-world large-scene MoCap dataset HPS~\cite{guzov2021human}. All methods are trained on the training sequences of the CMU, BMLrub, and HDM05 subsets of AMASS.
Tab.~\ref{tab:result_hps} and Fig.~\ref{fig:hps} present the numerical and visual results.
Our method achieves significantly better performance than existing methods. The results on the HPS dataset, encompassing motion data both in close proximity to the origin and at more distant points, demonstrate the robustness of our method across various spatial contexts. This is particularly noteworthy given that our model was exclusively trained on indoor MoCap data.

By adapting local representations, we also improved the performance of state-of-the-art methods on large-scale scenes. For each input window, we normalize the position of the head and hands to their positions in the first frame. This ensures that the network's input range is consistently observed during training, even when the user is far from the origin. This method significantly enhances the overall performance of state-of-the-art methods, particularly when the position is far from the origin. However, our method still outperforms these methods by a substantial margin.

We also show the comparisons of the number of parameters and inference time in Fig.~\ref{fig:flops}. Our method achieves significantly better performance with a much smaller model size and lower computational cost than recent methods.\\

\noindent\textbf{Pose estimation robustness relative to the distance from the origin.} We also analyze the robustness to location variations on the CMU, BMLrub, and HDM05 subsets. To do this, we add a constant positional offset of 0, 2, 5, 10, and 50\,m to each sequence to simulate a user performing a motion at different locations. We compare our method to AvatarPoser~\cite{jiang2022avatarposer}, AGRoL~\cite{du2023avatars}, and AvatarJLM~\cite{zheng2023realistic}, which use global input representations, and the classical KNN-based method CoolMoves~\cite{ahuja2021coolmoves}, and FinalIK~\cite{finalik}, which use local input representations.
We plot the position error against different offsets in Fig.~\ref{fig:result_amass}. 
While AvatarJLM achieved the best performance when no offset was applied, all prior learning-based methods, namely AvatarPoser, AGRoL, and AvatarJLM, taking global pose as input, experience a significant decrease in performance as the offset from the origin increases. 
In contrast, methods using local representations maintain stable performance.\\

\begin{table}[]

\begin{minipage}[t]{.42\linewidth}    
    \caption{Evaluation of shape-aware pose estimation on AMASS dataset. All metrics are measured in [cm].}
    \centering
    \small
    \setlength{\tabcolsep}{1.5pt}
    \begin{adjustbox}{width=\columnwidth,center}
    \begin{tabular}{@{}lcccccccc@{}}
        \toprule
        Strategies      & MPJPE & Vertex & Height & Arm & GP & FF  \\
        \midrule
      Mean Shape  &  6.36 & 6.74 & 7.67 & 7.42     & 3.87  & 5.38 \\  
    Ours 1 - DA + Calib.	  & 5.26  & 4.69  & \textbf{1.36}  & \textbf{1.24}   & \textbf{2.06} & 1.67 \\
    Ours 2 - Shape Est.	   &	 \textbf{4.79} & \textbf{4.08} & 1.78 & 1.66 & 2.31  & \textbf{1.64}      \\
        
        \bottomrule
    \end{tabular}
    \end{adjustbox}
    \label{tab:bodysize}
\end{minipage}
\hfill
\begin{minipage}[t]{.55\linewidth}    
  \centering
    \setlength{\tabcolsep}{1pt}
  \caption{Results of different methods under various field of views on AMASS dataset.}
    \begin{adjustbox}{width=\columnwidth,center}
    \begin{tabular}{l|rr|rr|rr}
    \toprule
     \multirow{2}{*}{Strategies} & \multicolumn{2}{c|}{FoV = 180\degree} & \multicolumn{2}{c|}{FoV = 120\degree} & \multicolumn{2}{c}{FoV = 90\degree} \\ 
      & MPJPE & MPJVE & MPJPE & MPJVE & MPJPE & MPJVE\\ \midrule
    \multicolumn{1}{l|}{Full Visibility~\cite{jiang2022avatarposer}} & 24.75 & 183.84 & 38.99 & 144.42 & 41.24 & 95.66 \\ 
    \multicolumn{1}{l|}{Random Masking~\cite{aliakbarian2022flag, du2023avatars}} & 7.09 & 49.91 & 13.29 & 64.09 & 14.84 & 58.33\\ 
    \multicolumn{1}{l|}{Improved RM} & 6.52 & 47.50 & 11.88 & 57.44 & 12.83 & 52.98\\     
    \multicolumn{1}{l|}{Ours}  & \textbf{5.31} & \textbf{39.69} & \textbf{6.07} & \textbf{46.01} & \textbf{6.60} & \textbf{48.25} \\
        \bottomrule
  \end{tabular}
\end{adjustbox}
  \label{tab:fov}
\end{minipage}

\end{table}

\begin{table}[h]
\begin{minipage}[t]{.42\linewidth}    

    \centering
    \caption{Ablation study of different global motion decomposition methods using the AMASS dataset.}
    \small
    \setlength{\tabcolsep}{1pt}

\begin{adjustbox}{width=\columnwidth,center}

    \begin{tabular}{@{}l rrr@{}}
        \toprule

        Strategies          & MPJPE & MPJVE \\
        \midrule
Mean Norm. (all features)    	            		&	6.25	&	42.69	\\
Mean Norm. (horiz. + vert. pos.)      	            		&	6.24	&	42.75	\\ 
Mean Norm. (horiz. pos.)      	            		&	6.25	&	42.87	\\ 
Spatial Norm. (horiz. + vert. pos.) &	4.96	&	29.59	\\
Spatial Norm. (horiz. pos.) &	4.45	&	27.56	\\
$\text{Temporal Norm.}$ & 4.58	& 28.01\\
Ours&	\textbf{4.14}	&	\textbf{25.95}	\\

        \bottomrule
    \end{tabular}
    \end{adjustbox}
    \label{tab:gil_ablation}

\end{minipage}
\hfill
\begin{minipage}[t]{.55\linewidth}


    \caption{Ablation study of the SlowFast design on AMASS dataset. Our design choice captures longer time series without introducing additional computational costs.}
    \centering
    \setlength{\tabcolsep}{5.0pt}

 \begin{adjustbox}{width=\columnwidth,center}
    \begin{tabular}{@{}l cccc@{}}
        \toprule
        Strategies          & MPJPE & MPJVE & FLOPs & \#Parameters\\
        \midrule
length 40         &	4.36	&	28.12	&	0.33G& 4.12M	\\
length 80         &	\textbf{4.11}	&	29.27	&	0.65G& 4.12M	\\
length 80 , s=2     &	4.13	&	30.02	&	0.33G& 4.12M	\\
Ours	        &	4.14	&	\textbf{25.95}	& 0.33G& 4.12M	\\
        \bottomrule
    \end{tabular}
\end{adjustbox}
    \label{tab:ablation_slowfast}

\end{minipage}
\end{table}

\noindent\textbf{Results of outside-the-FoV pose estimation.}
We evaluated various strategies for scenarios where hands are tracked by a headset and may go out of the cameras' FoV.
To simulate real-world scenarios, we considered various angles of available FoV: 180\degree (in fisheye cameras), 120\degree (in Quest~2), 90\degree (in Hololens~2).
We tested the results on a model trained on hands with full visibility, denoted as `Full visibility', on fine-tuned models with random hand masking using a probability $p = 0.2$ as proposed in FLAG~\cite{aliakbarian2022flag} (denoted as `Random Masking' or `RM'), and on our realistic FoV modeling (denoted as `Ours').

\begin{figure}[h]
    \centering
    \begin{minipage}{0.48\textwidth}
        \centering
        \includegraphics[width=\linewidth]{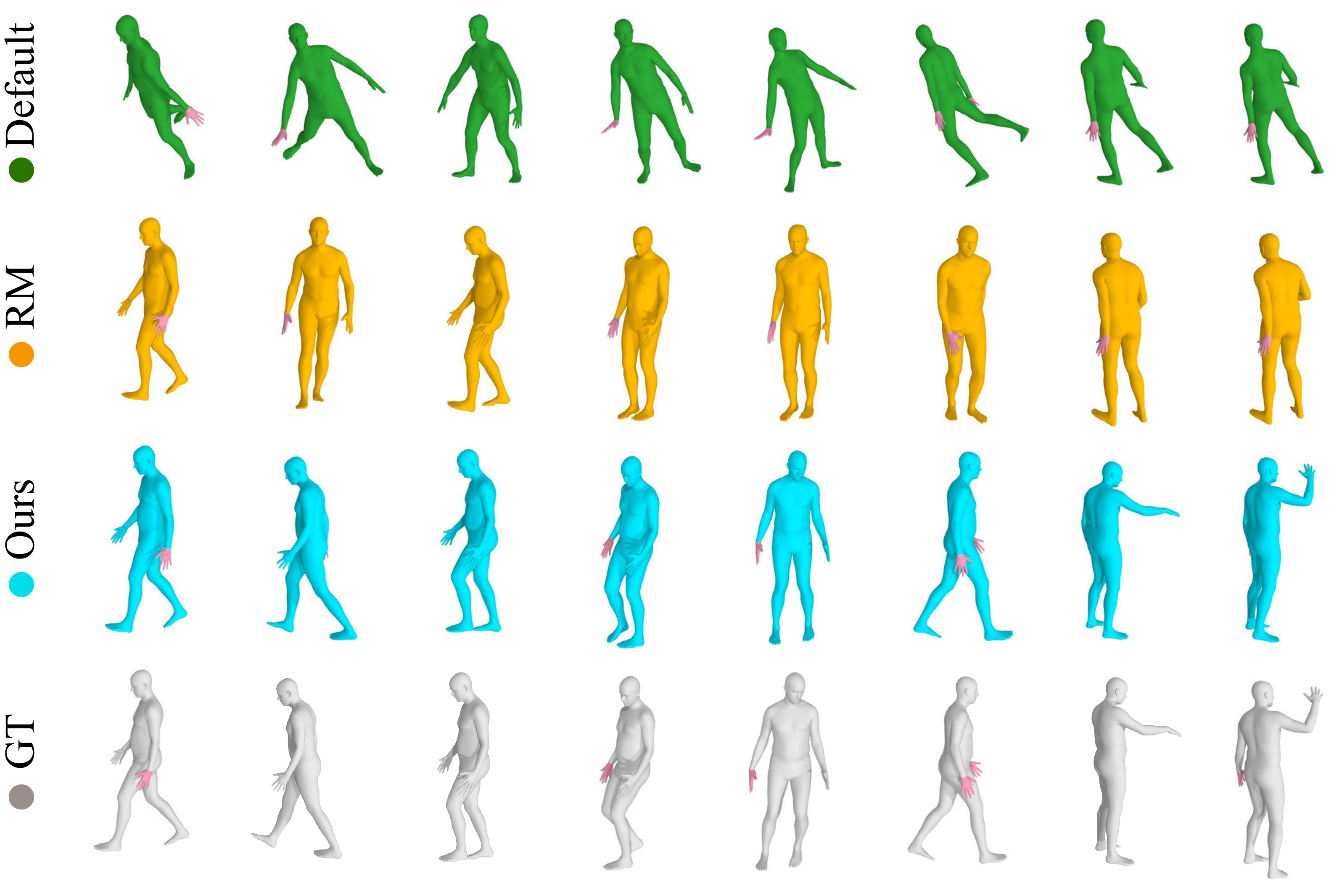}
        \caption{Visual comparisons between methods for the scenario where the hands can go out of FoV. Hands outside FoV are rendered in \textcolor{mypink}{\CIRCLE}.
        }
        \label{fig:fov_compare}
    \end{minipage}
    \hfill
    \begin{minipage}{0.48\textwidth}
        \centering
        \includegraphics[width=\linewidth]{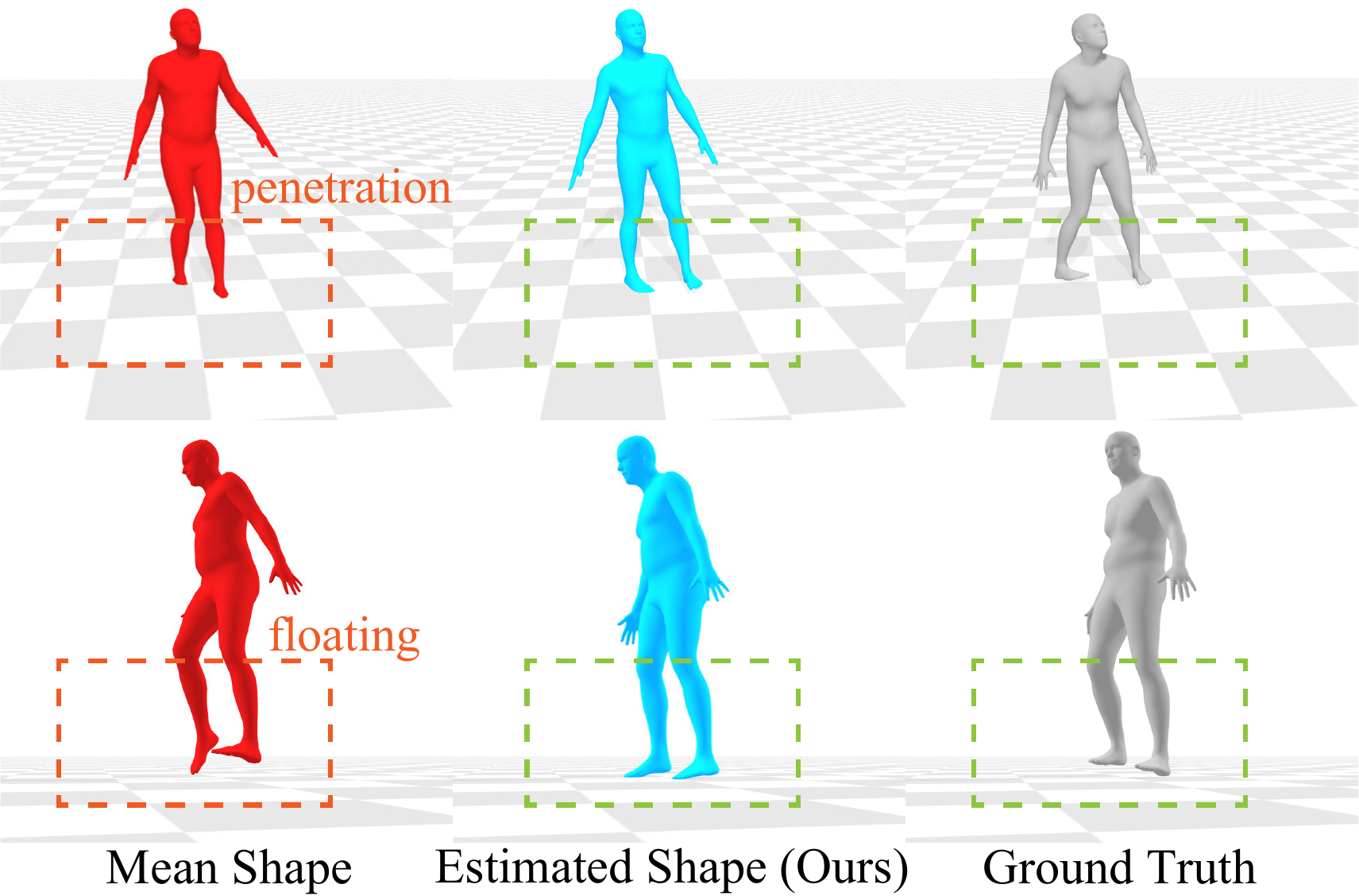}
        \caption{Visual comparison between results using the mean shape and estimated shape. Our method reduces penetration and floating.}
        \label{fig:shape}
    \end{minipage}
\end{figure}

Tab.~\ref{tab:fov} and Fig.~\ref{fig:fov_compare} present the numerical and visual results of models trained with different strategies. When testing the performance on various FoVs using the default model that assumes hands are always visible during training, we observe two main trends. First, as the FoV becomes smaller, the position error MPJPE increases. This is intuitive since a smaller FoV means there is more chance that hands are outside the FoV, rendering accurate predictions more challenging.
Second, with a smaller FoV, the velocity error MPJVE initially increases and then decreases. This trend can be explained with the strong discontinuity in predictions for FoVs of 180\degree or 120\degree, switching between hands leaving and reentering the FoV. With even smaller FoVs, hands are mostly or even always outside the FoV, leading to smoother but less accurate predictions. 

We also improve the random masking strategy. While random hand masking can improve results, our realistic FoV modeling strategy sets the visibility status based on the actual position of the hand relative to the head and captures the real temporal dependencies of hand visibility. It thus achieves the best performance for both position accuracy and smoothness.

\noindent\textbf{Results of shape-aware pose estimation.}
We evaluate the performance of the shape-aware pose estimation on the same test data as in Fig.~\ref{fig:result_amass} from AMASS but with the true shape parameters $\mathbf{\beta}$.
This test set includes over 175~subjects with heights ranging from 145 to 207\,cm. As listed in Tab.~\ref{tab:bodysize}, the model trained using the mean body shape achieved a mere 6.36\,cm in MPJPE.
The average error for body dimensions such as height and arm length exceeds 7\,cm.
These discrepancies arise from an inaccurate shape representation, which leads to issues like ground penetration (GP) and floating feet (FF) as shown in Fig.~\ref{fig:shape}. 
Metrics are in centimeters~[cm].

Conversely, as our first solution, data augmentation (DA) with ground truth body shapes and subsequently re-scaling the standardized model output by a body size factor obtained via T-pose calibration reduced the MPJPE to 5.26\,cm.
It also considerably enhanced performance across various metrics for body sizes and motion artifacts.
Our calibration-free shape prediction approach showed further improvements in both MPJPE and mean vertex error, while delivering comparable outcomes in metrics for body size and motion artifacts.
Calibration works slightly better for arm length and height as they are directly measured. 

\subsection{Ablation Studies}
We conduct thorough ablation studies to show the effectiveness of each proposed component. The different approaches for shape-aware and FoV-aware pose estimation have already been discussed in Tab.~\ref{tab:bodysize} and Tab.~\ref{tab:fov}. 

Tab.~\ref{tab:gil_ablation} lists our ablation studies on global motion decomposition.
Mean normalization refers to removing the mean value of each feature across the temporal window.
Alternatively, we substract only the mean of the vertical or horizontal joint translations.
Spatial normalization refers to subtracting the horizontal and/or vertical head translation from all joint positions per frame.
The results indicate that retaining vertical position information leads to better predictions.
We also experiment with applying only temporal normalization.
Combining our proposed temporal and spatial normalization achieves the best performance. 

Tab.~\ref{tab:ablation_slowfast} shows ablation studies for the SlowFast design. We compare our approach with methods that utilize signals of varying frame lengths: 40, 80, or the original 80 samples downsampled by a factor of 2. The results demonstrate that our design benefits from extended temporal context without significantly increasing the model size or computational cost.

\subsection{Demos on Commodity Mixed Reality Devices}
\begin{figure}[t]
\centering\includegraphics[width=\linewidth]{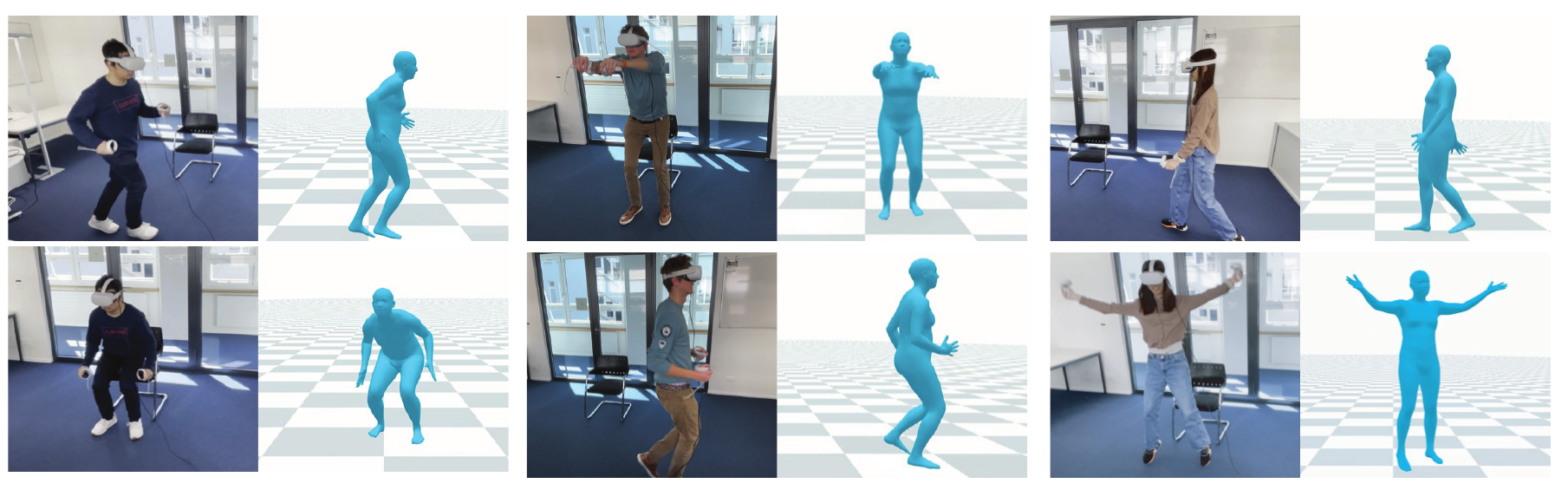}
    \caption{\method's visual demos on real data from Quest~2 with different participants.}
    \label{fig:demo}
\end{figure}

To assess \method's robustness on actual real-world data from an end-user device, we ran it on live recordings from participants with diverse body shapes, wearing Meta Quest~2 headsets, recorded by Velt~\cite{fender2018velt}.
Fig.~\ref{fig:demo} shows the results.

\subsection{Limitations and Discussions}
\label{sec:limitations}
While our method has demonstrated generalization ability for the challenging problems in HMD-based egocentric pose estimation, several limitations remain. First, since we encode vertical global position as an input feature, we assume that the user is moving on the same floor. Therefore, the origin system needs to be redefined when moving to another floor.
Second, we address real-world inside-out tracking problems in this paper using a straightforward Transformer as our backbone model. Future work could investigate more elaborate models for this task.
Third, our method has no post-processing steps. This could however further improve estimation accuracy or enhance physical plausibility. 

\section{Conclusion}

We have proposed \method, a novel systematic approach for 3D full-body pose estimation based solely on the tracking information available on contemporary Mixed Reality head-mounted devices.
We address the challenges faced by existing efforts using such platforms, specifically scaling robust estimations to arbitrary real-world settings, handling hands as input even when they are outside the cameras' field of view, and robustness to varying body dimensions.
Our experiments showed that \method achieves new state-of-the-art performance for accurate motion estimation under these challenging circumstances by combining our novel global motion decomposition, SlowFast fusion strategy, robust field-of-view modeling, and shape-aware pose estimation method.
We believe that our proposed strategies can significantly contribute to the advancement of 3D full-body pose estimation and its integration into various AR/VR/MR applications.\\

\noindent\textbf{Acknowledgments}\\[.2em]
We sincerely thank Andreas Fender for his help with data recording, testing, and manuscript proofreading.

\bibliographystyle{splncs04}
\bibliography{literature}
\clearpage

\end{document}